# An Improved Rapidly Exploring Random Tree Algorithm for Path Planning in Configuration Spaces with Narrow Channels

Mathew Mithra Noel[1], Akshay Chawla[2]

School of Electrical Engineering, Vellore Institute of Technology, Vellore, India[1]
Carnegie Mellon University, USA[2]

mathew.m@vit.ac.in[1]
akshaych@alumni.cmu.edu[2]

## ABSTRACT

Rapidly-exploring Random Tree (RRT) algorithms have been applied successfully to challenging robot motion planning and under-actuated nonlinear control problems. However a fundamental limitation of the RRT approach is the slow convergence in configuration spaces with narrow channels because of the small probability of generating test points inside narrow channels. This paper presents an improved RRT algorithm that takes advantage of narrow channels between the initial and goal states to find shorter paths by improving the exploration of narrow regions in the configuration space. The proposed algorithm detects the presence of narrow channel by checking for collision of neighborhood points with the infeasible set and attempts to add points within narrow channels with a predetermined bias. This approach is compared with the classical RRT and its variants on a variety of benchmark planning problems. Simulation results indicate that the algorithm presented in this paper computes a significantly shorter path in spaces with narrow channels.

**Keywords:** Rapidly-exploring Random Tree (RRT); Path Planning; Robotics; Motion Planning; Monte Carlo method; Constrained Nonlinear Control

# I. INTRODUCTION

*1.1 Introduction to path planning*

The problem of finding a path between a start state and a goal state in a geometrically complex non-convex abstract configuration space is fundamental to the solution of a wide array of problems in robotics and nonlinear control. Applications include robot path and motion planning [1], autonomous urban driving [2], control of under-actuated nonlinear systems [3] and maneuvering of aerial vehicles like Quadcopters in confined spaces [4]-[6]. Applications of path planners are not limited to the field of robotics and control; bio-engineers use path planners for molecule disassembly [7] and computer graphics engineers use planners to simulate motions of virtual objects in complex environments [8], [9].

Path planning algorithms can be classified as Graph based, Tree based and Hybrid methods. The Probabilistic Road Map (PRM) algorithm [10] is a Graph-based algorithm. PRMs operate by generating random points in the configuration space and then connecting nearest neighbors in the feasible set with a local planner. The starting and goal configurations are added in, and a graph search algorithm is applied to the resulting graph to determine a path between the starting and goal configurations. However the problem of finding a path connecting nearest neighbors is a challenging boundary value problem in nonlinear control applications. Hence the success of the PRM approach depends on the existence of a good local planning algorithm. Thus path planning algorithms that do not require a local planner or involve solution of complex nonlinear boundary value problems are of interest.

An efficient Tree based path planning algorithm that does not require solution of complex boundary value problems based on the Rapidly-exploring Random Tree (RRT) approach was developed by S. M. LaValle and J. J. Kuffner [11]. The essential idea behind the RRT approach is the construction of a space-filling tree in the configuration space. Consider the classical problem of guiding a mobile robot from a start point to a goal point in a 2D planar region with obstacles. In this case the configuration space is the set of (x,y) positions of the robot. The set of positions not intersected by any obstacle in referred to as $C_{free}$. Path planning is the problem of finding a path between a starting point $x_{start}$ and goal $x_{goal}$ inside $C_{free}$. In general the configuration space C might represent the Joint Space of a robot or the State Space of nonlinear system and is usually non-convex with complex geometry. An exhaustive search for a path between the start and goal states is computationally intractable and hence RRTs employ a random sampling based approach to explore the configuration space and find a collision free path. Sampling based approaches take advantage of efficient algorithms to generate random samples in the configuration space and check for collisions with the obstacle region. RRTs have been successfully applied to challenging path planning problems like the problem of finding a collision free path in real time for agile maneuvers of robots in complex dynamic environments [12], [13], [14]. Hybrid approaches combine the global coverage of PRMs with the local fast coverage of RRTs to provide shorter paths efficiently [15]. A wide variety of other sampling based path planning algorithms are discussed in [16].

RRTs operate by building a space filling tree in $C_{free}$ between the start and the goal configurations/states. The RRT starts with the initial state as the first node in the space filling tree T and attempts to add a random point $x_{rand}$ in $C_{free}$ every iteration. If the random sample is not in $C_{free}$ it is discarded else the tree is grown by adding a branch between the random sample $x_{rand}$ and the nearest node in T. When sufficient number of points have been added to T the RRT either finds a path or reports failure. Since the RRT uniformly samples the configuration space it is not biased to grow towards the goal and hence can suffer from slow convergence [17], [18]. Heuristics such as goal bias in sampling [18] and bidirectional growth [19] can significantly enhance the performance of RRTs.

Despite the wide application of RRTs a fundamental limitation of the RRT approach is the slow convergence in spaces with narrow channels between the start and the goal states. Narrow channels in the configuration space are difficult to explore because the probability of generating random samples inside narrow channels is very small [18], [20]. Path planning problems with narrow channels between Start and Goal states arise in applications involving congested and confined spaces. Examples include path planning for small aerial robots (like Quadcopters) inside buildings and autonomous driving on congested roads. Thus the problem of path planning in spaces with narrow channels is of interest.

In this paper an improved RRT that computes shorter paths by attempting to generate samples inside narrow channels with a predetermined bias is proposed and compared with existing RRTs from literature. The paper is organized as follows: In section 1, the mathematical formulation of the path planning problem and a variety of existing RRT algorithms is presented.

In section 2, an improved RRT that takes advantages of paths in narrow channels is presented. Finally, the performance of various RRTs on challenging benchmark planning problems is presented (Section 3) followed by a discussion of the results (Section 4).

*1.2 Problem formulation*

The terminology used in the paper is defined below:

1. **State/Configuration space (C):** A Metric Space denoted by C. The dimensions of the state space is determined by the number of degrees of freedom in robot path planning problems and by the number of state variables in nonlinear control problems.
2. **Tree (T):** The RRT is represented by a Tree data structure. Mathematically a Tree is a connected Graph with no cycles (no closed walks).
3. **Start** ($x_{init}$) **and End** ($x_{goal}$) **configurations:** The start configuration and the end configuration define the goals of the planner.
4. **Obstacle region** ($C_{obs} \subset C$)**:** It is the subset of points in the configuration state that represent physical obstacles or unattainable configurations.
5. **Collision detection function *f*:** It is a function $f : C \to \{True, False\}$ that returns *True* when the configuration *x* falls in the obstacle region and *False* otherwise.
6. **Free space** ($C_{free} \subset C$)**:** The free space is a subset of C that represents collision free configurations.
7. **Metric ($\rho$):** Is a non-negative real valued function $\rho : C \times C \to R$ that denotes the distance between any two arbitrary configurations and satisfies the axioms of a Metric Space. In this paper the Euclidean distance between points is used as the metric.

## II. PATH PLANNING ALGORITHMS

This section describes the first RRT algorithm, **RRTBasic** introduced by S. M. LaValle [11] and its improvements presented in [18]. The standard RRT operates by incrementally growing tree data structure rooted at the start configuration $x_{init}$. At every iteration, a random sample in free space is generated by the algorithm (*x*) and an attempt is made to add this configuration to the tree. If the distance between the new configuration and the nearest node in the tree is greater than a threshold $\varepsilon$, another configuration $x_{new}$ is chosen along the line connecting $x_{near}$ and $x$ such that $\rho(x_{near}, x_{new}) \leq \varepsilon$. Thus, the RRT algorithm uses random samples from the configuration space to pull the tree towards unexplored areas of the state space. This process is continued till the goal configuration is sufficiently close to the tree. The algorithm then adds the goal configuration to the tree and returns the completed RRT. The constructed tree has no closed paths or self-loops and hence there is exactly one path from $x_{init}$ to $x_{goal}$. Thus the unique path between $x_{init}$ and $x_{goal}$ can be computed by traversing the tree in the reverse direction starting at the leaf node $x_{goal}$ and going up to the root node $x_{init}$.

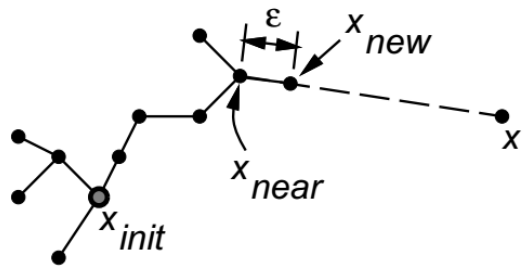

Figure 1. The *EXTEND* Function of the **RRTBasic** algorithm.

---

***Build_RRTBasic(xinit, xgoal)***

1. T.init($x_{init}$)
2. **for** k=1 to K, **do**
    2.1. $x_{rand}$ ← **RANDOM_STATE**()
    2.2. **EXTEND**(T, $x_{rand}$)
    2.3. $x_{near}$ = **NEAREST_NEIGHBOUR**($x_{goal}$, T)
    2.4. **if** $\rho(x_{near}, x_{goal}) \leq \varepsilon$ **do**
        2.4.1. rc = **EXTEND**(T, $x_{goal}$)
        2.4.2. **if** rc == *Reached* **do**
            2.4.2.1. Return T
3. Return T

---

Table 1. The **RRTBasic** algorithm

*EXTEND(T, x)*

1. $x_{near} \leftarrow$ *NEAREST_NEIGHBOUR(x, T)*
2. $x_{new} \leftarrow$ *NEW_STATE(x, $x_{near}$, ε)*
3. **if NOT** *COLLISION_CHECK($x_{near}$)* **do**
    3.1. T.add_vertex($x_{new}$)
    3.2. T.add_edge($x_{near}$, $x_{new}$)
    3.3. **if** $x_{new} = x$ **do**
        3.3.1. Return *Reached*
    3.4. **else**
        3.4.1. Return *Advanced*
4. **else**
    4.1. Return *Trapped*

Table 2. The *EXTEND* function used in the **RRTBasic** algorithm.

| Function | Definition |
|---|---|
| *RANDOM_STATE()* | Returns a random configuration in free space. |
| *NARROW_STATE()* | Returns a configuration close to or within a narrow channel. |
| *NEAREST_NEIGHBOUR(x, T)* | Returns the nearest node in the Tree *T* to the configuration *x*. |
| $\rho(x_1, x_2)$ | Euclidean distance between configurations $x_1$ and $x_2$. |
| *COLLISION_CHECK(x)* | Returns true if the configuration *x* falls in the obstacle region in state space. |

Table 3. Summary of functions used by RRTs

In order to improve the speed of convergence to the goal state, variants of the **RRTBasic** algorithm that introduce a small bias towards the goal state were proposed. **RRTGoalBias** [18] is a variant wherein $x_{rand}$ is determined by tossing a biased coin (probability of a head p is taken to be a small number like 0.1). If the coin toss yields a head, then $x_{goal}$ is returned, else a random configuration is returned. Thus in **RRTGoalBias** an attempt is made to add the goal state directly with a small probability instead of a random generated state every iteration. Another variant, **RRTGoalZoom** [18] also tosses a biased coin but instead of returning $x_{goal}$, it returns a configuration between $x_{goal}$, and the closest node in the tree. A visual representation of the tree expansion for the algorithms mentioned above is shown in figures 2 - 4. An alternative approach to increase convergence is by using bi-directional planners; where two RRTs are grown, one rooted at $x_{init}$ and one rooted at $x_{goal}$.

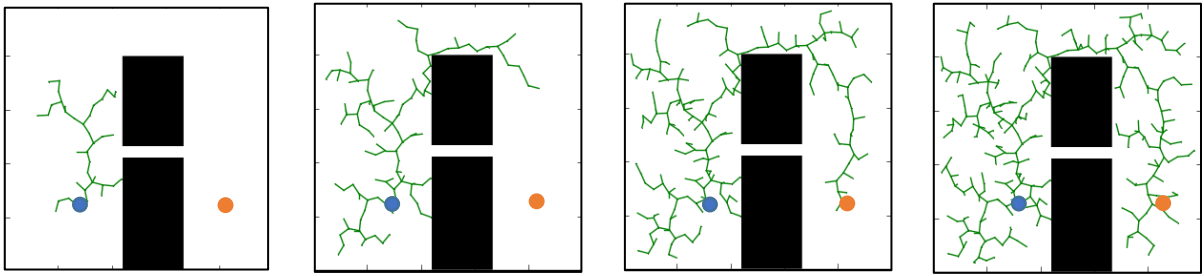

Figure 2. Growth of the RRT from the starting state (blue point) towards the goal state (orange) for the **RRTBasic** algorithm.

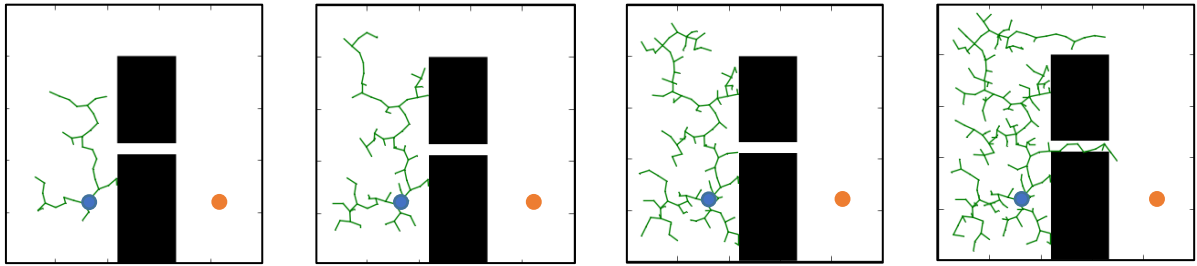

Figure 3. Growth of the RRT from the starting state (blue point) towards the goal state (orange) for the **RRTGoalBias** algorithm.

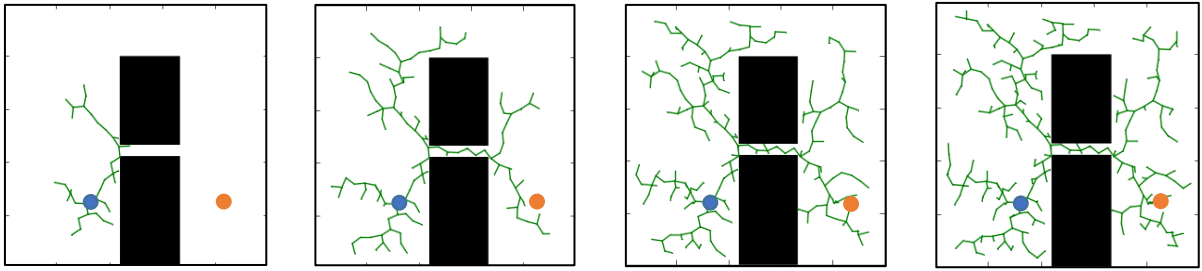

Figure 4. Growth of the RRT from the starting state (blue point) towards the goal state (orange) for the **RRTGoalZoom** algorithm.

## III. NARROW CHANNEL RRT (**NCRRT**) ALGORITHM

The NCRRT algorithm attempts to improve the exploration of narrow channels in the configuration/state space by incorporating a narrow channel bias. In order to decide whether a random configuration $x_{rand}$ is in a narrow space, the algorithm samples a cluster of neighboring configurations around $x_{rand}$. This cluster of points is chosen to be within a neighborhood of radius $\lambda$ of $x_{rand}$. These neighboring configurations are then checked for collision. If the fraction of points that collide with the obstacle region exceeds a user set threshold $\sigma$, $x_{rand}$ is assumed to lie inside a narrow channel. Figure 5 shows the narrow channel check being performed for the red point. The green points corresponds to configurations in free space and yellow points to infeasible configurations.

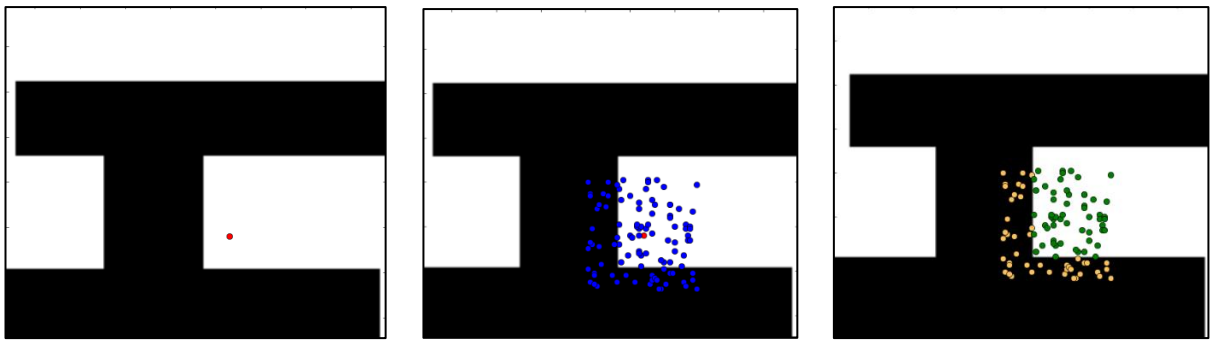

Figure 5. Cluster sampling

In order to incorporate a narrow channel bias into a standard RRT model, the **BUILD_RRTBasic** function from Table 1 is modified and presented in Table 4.

---

*Build_NCRRT*($x_{init}$)

---

4. T.init($x_{init}$)
5. **for** k=1 to K, **do**
    5.1. **if** mod(k, $\alpha$)==0 **do**
        5.1.1. $x_{rand}$ ← *NARROW_STATE*()
    5.2. **else**
        5.2.1. $x_{rand}$ ← *RANDOM_STATE*()
    5.3. *EXTEND*(T, $x_{rand}$)
    5.4. $x_{near}$ = *NEAREST_NEIGHBOUR*($x_{goal}$, T)
    5.5. **if** $\rho(x_{near}, x_{goal}) \leq \varepsilon$ **do**
        5.5.1. rc = *EXTEND*(T, $x_{goal}$)
        5.5.2. **if** rc == *Reached* **do**
            5.5.2.1. Return T
6. Return T

---

Table 4. The **NCRRT** Algorithm

The parameter $\alpha$ controls the aggressiveness of the narrow channel bias. Decreasing the value of $\alpha$ will increase the number of times a configuration in a narrow space will be chosen over a random configuration. This is important because having a very low value of $\alpha$ will decrease the ability of the planner to find paths through broad spaces in the environment and thus increase the iterations required to reach the goal when no narrow channels are present in the environment. The function *NARROW_STATE* utilizes the aforementioned sampling technique (Figure 5) to generate a configuration. Figure 6 shows a sample run of the **NCRRT** algorithm, it can be clearly seen that in comparison to figures 2 – 4, the **NCRRT** algorithm actively tries to add configurations that lie within narrow channels to the tree.

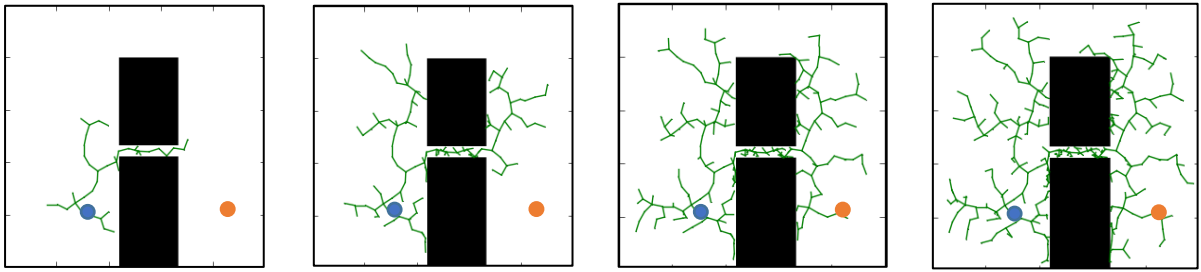

Figure 6. Growth of **NCRRT** from the start configuration (blue) towards the goal configuration (orange).

## IV. RESULTS

### *4.1 Methodology*

**NCRRT** was compared to a variety of RRTs such as **RRTBasic, RRTGoalBias and RRTGoalZoom.** The comparison methodology involved testing these algorithms on 4 benchmark path planning scenarios with the same initial and goal configurations, as shown in figure 7. The planning scenarios considered are such that an algorithm can either find a shorter path through narrow channels or a longer path through broad areas. In the following figures white areas represent configurations or positions that can be attained by a robot while the black areas are the obstacle regions. Blue dots denote the start configuration and yellow dots denote the goal configuration. Scenarios 1 and 2 were inspired by the path planning scenarios presented in a key paper on RRTs [21] and hence reflect good test cases for tree based planners.

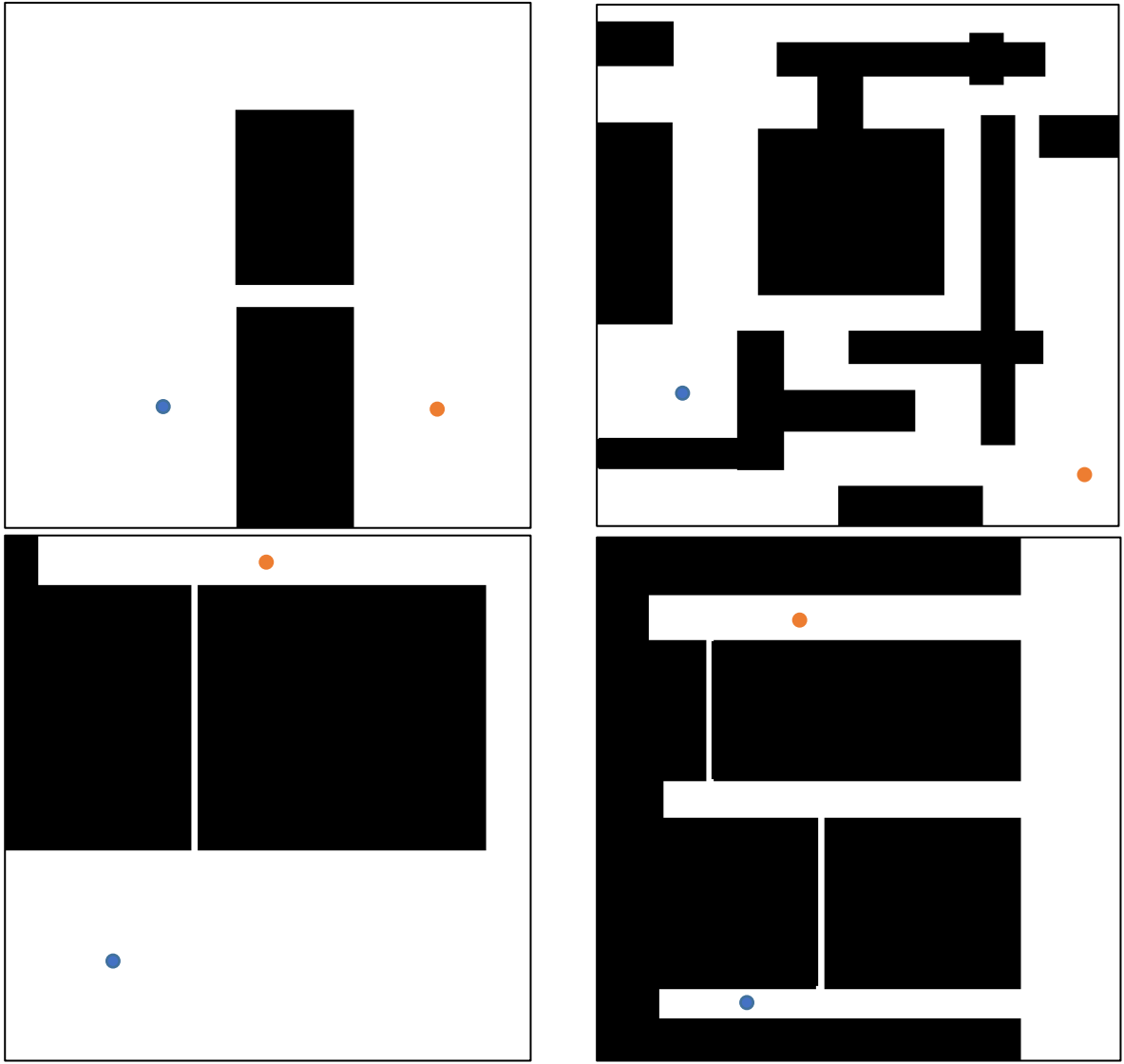

Figure 7. Four benchmark path planning scenarios with start and goal configurations.

Since RRTs are stochastic in nature the average performance over multiple independent runs must be considered. In this paper the performance of each algorithm on benchmark path planning problems over 100 independent runs is considered. Additionally, an upper limit of 1500 iterations was placed on the algorithm so that a failure is reported if the iterations required to reach the goal exceed 1500. For each trial run, the path found, path length and iterations required were recorded. The algorithms were run using the parameters specified in Table 5. The performance of different algorithm are judged based on the following performance metrics:

- Length of the path found
- Fraction of narrow channel paths generated
- Average execution time

In order to compare the path lengths, a 30 bin histogram of the results obtained during the 100 trials was generated. This showed the distribution of short and long paths taken by each of the algorithms. Based on the histogram, a threshold was set that distinguished between a short path and a long path. The number of short and long paths were then counted and compared for each algorithm.

| Algorithm | Parameters |
|---|---|
| **RRTBasic** | $\varepsilon = 20$, K=1500 |
| **RRTGoalBias** | $\varepsilon = 20$, p = 0.9, K=1500 |
| **RRTGoalZoom** | $\varepsilon = 20$, p = 0.9, K=1500 |
| **NCRRT** | $\varepsilon = 20$, p = 0.9, $\sigma = 40$, $\alpha = 3$, $\lambda=20$, K=1500 |

Table 5. Parameter setting for different RRTs

## 4.2 Length of path found

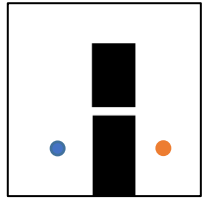

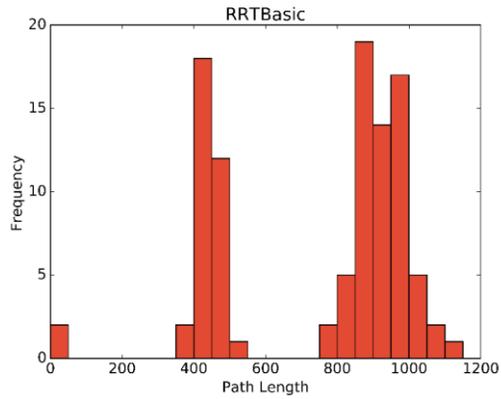
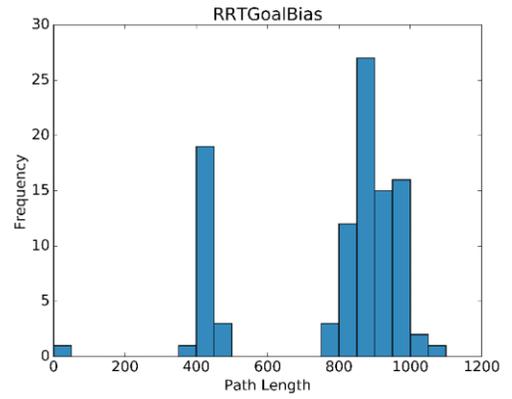
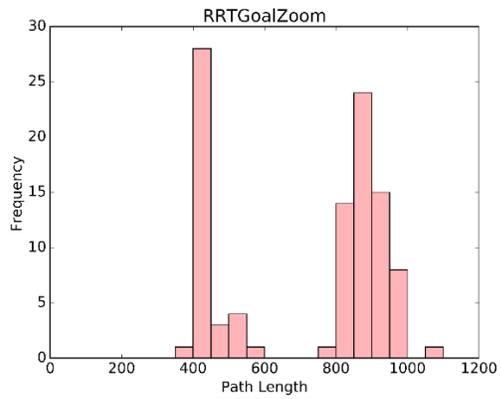
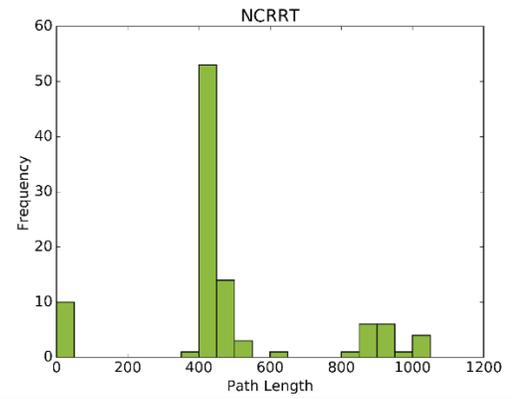

Figure 8. Histograms of final path lengths found by RRT variants for benchmark path planning scenario 1.

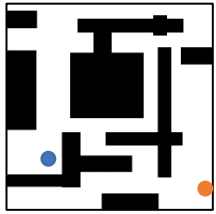

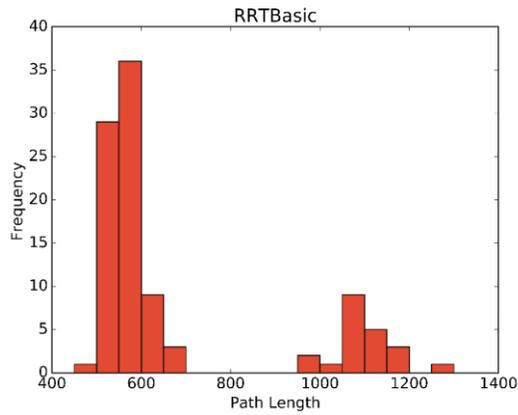
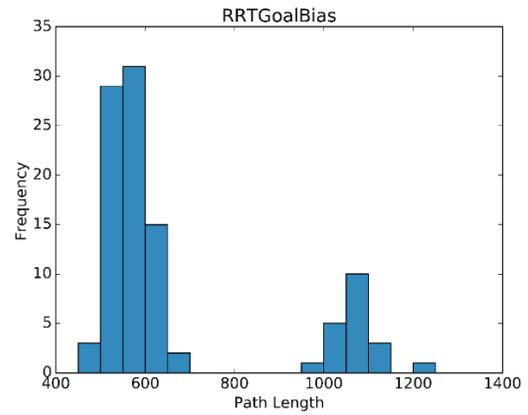
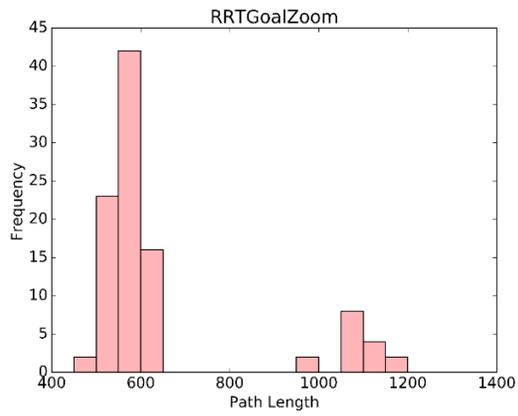
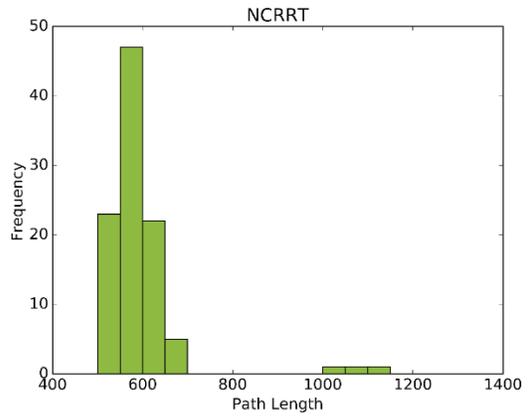

Figure 9. Histograms of final path lengths found by RRT variants for benchmark path planning scenario 2

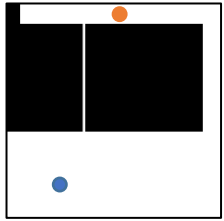

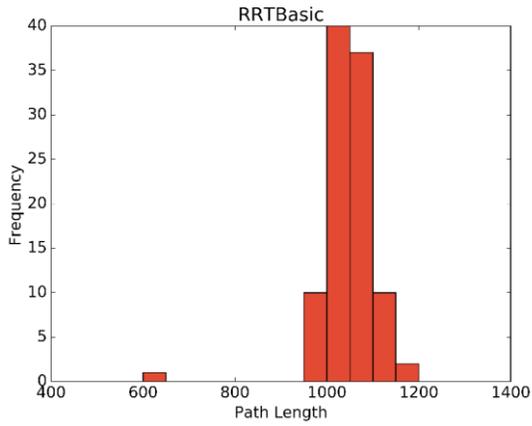
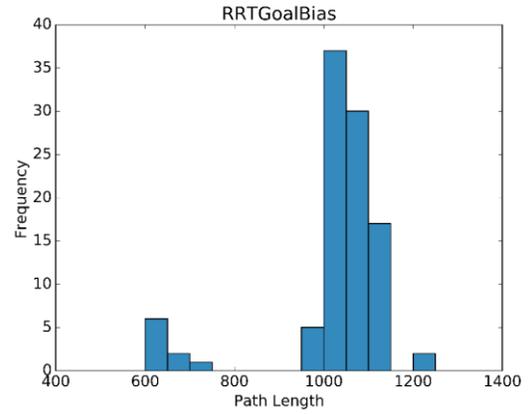
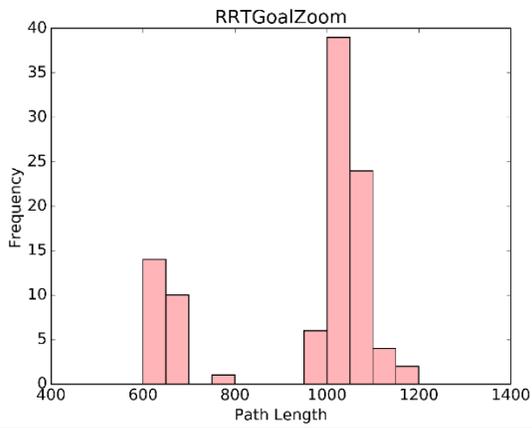
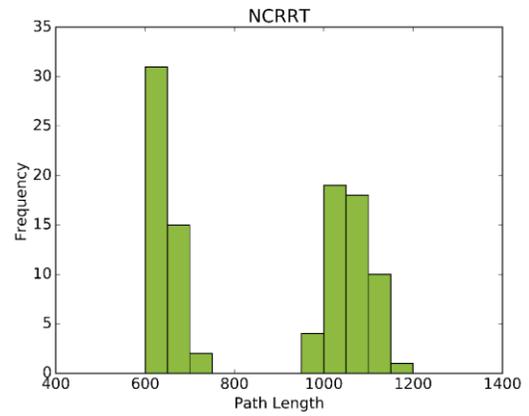

Figure 10. Histograms of final path lengths found by RRT variants for benchmark path planning 3

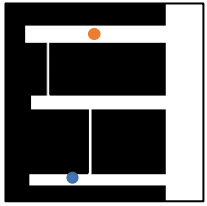

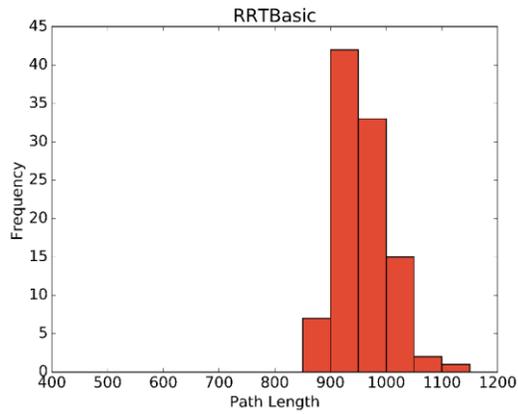
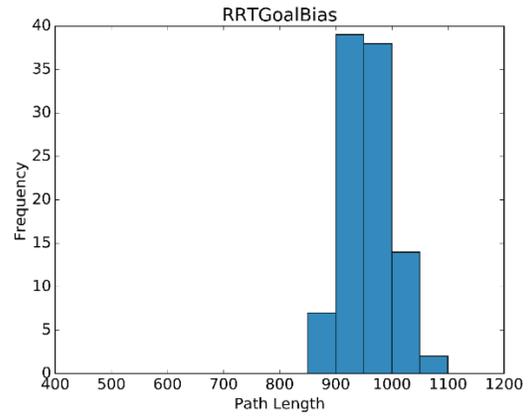
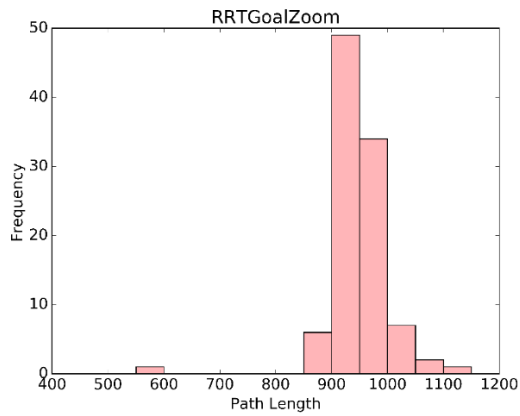
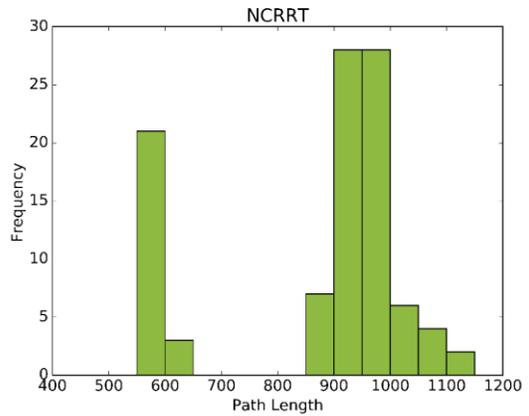

Figure 11. Histograms of final path lengths found by RRT variants for benchmark path planning 4

| | 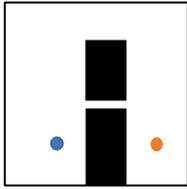 | 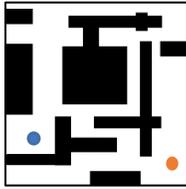 | 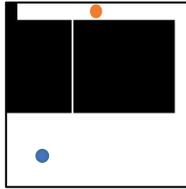 | 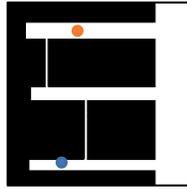 |
|---|---|---|---|---|
| **RRTBasic**    | μ=749, σ=256 | μ=669, σ=229 | μ=1049, σ=57  | μ=959, σ=44  |
| **RRTGoalBias** | μ=781, σ=220 | μ=665, σ=209 | μ=1026, σ=124 | μ=956, σ=43  |
| **RRTGoalZoom** | μ=726, σ=222 | μ=645, σ=204 | μ=950, σ=172  | μ=945, σ=54  |
| **NCRRT**       | μ=483, σ=247 | μ=592, σ=95  | μ=861, σ=213  | μ=860, σ=187 |

Table 6. Mean and standard deviation of path lengths found by RRT variants

*4.3 Fraction of narrow channel paths generated*

Table 6 shows the fraction of shorter paths found by each algorithm for different benchmark planning problems.

| | 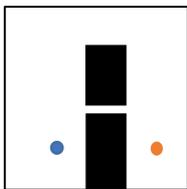 | 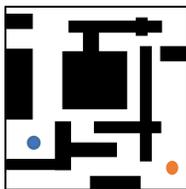 | 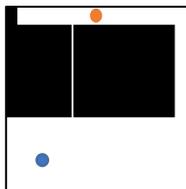 | 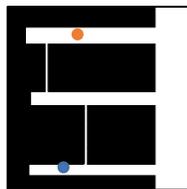 |
|---|---|---|---|---|
| **RRTBasic**    | 0.35 | 0.79 | 0.01 | 0.00 |
| **RRTGoalBias** | 0.24 | 0.80 | 0.09 | 0.00 |
| **RRTGoalZoom** | 0.37 | 0.84 | 0.25 | 0.01 |
| **NCRRT**       | 0.82 | 0.97 | 0.48 | 0.25 |

Table 7. Fraction of shorter paths found by RRT variants

*4.4 Average Execution Time*

Each of the algorithms, RRTBasic, RRTGoalBias, RRTGoalZoom and NCRRT were tested with the benchmark problems shown in Figure 7. The average execution time over 30 independent trials presented in Table 8.

|  | 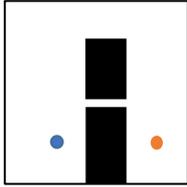 | 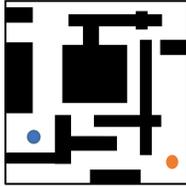 | 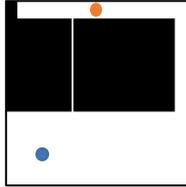 | 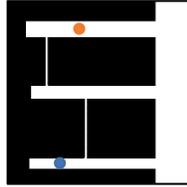 |
|---|---|---|---|---|
| **RRTBasic** | 0.620 | 5.431 | 6.832 | 0.75 |
| **RRTGoalBias** | 0.498 | 3.910 | 5.775 | 0.58 |
| **RRTGoalZoom** | 0.539 | 3.341 | 4.304 | 0.63 |
| **RRTNC** | 0.985 | 3.651 | 8.321 | 1.12 |

Table 8. Average execution time of RRT variants

## V. DISCUSSION OF THE RESULTS

Histograms in figures 8 – 11 show the performance of the various RRT based algorithms on a set of four benchmark path planning scenarios. Each 30-bin histogram represents the lengths of the path found in 100 trials of the respective algorithm-scenario pair. The histograms are color coded across maps for ease of comparison. The path-length data in the histograms is condensed into a tuple (mean, standard deviation) and presented in Table 6.

Table 7 shows that NCRRT has a higher chance of finding a shorter path compared to other algorithms. The benchmark problems considered are such that the shortest path is always through a narrow channel, while a longer but easy to find route through broad areas, also exists. Inspection of the histograms shows a prominent peak for shorter paths. Thus the mean path length is observed to be significantly smaller in case of **NCRRT**.

For scenario 4, where the shorter path passes through multiple narrow openings, **NCRRT** is able to find these narrow openings while **RRTBasic**, **RRTGoalBias** and **RRTGoalZoom** are unable to find a shorter path as indicated by the histograms.

For scenarios 1, 3 and 4 the difference between execution times for **NCRRT** and **RRTBasic** is small (Table 8), although **NCRRT** takes more time than **RRTGoalBias** and **RRTGoalZoom**. For scenario 2 **NCRRT** performs better than **RRTBasic** and **RRTGoalBias** in terms of both execution time and fraction of shorter paths. Thus the **NCCRT** algorithm proposed in this paper finds shorter paths compared to other RRTs without a significant increase in execution time.

## VI. CONCLUSION

The RRT approach to path planning has been used to solve an impressive variety of challenging robot path planning and nonlinear control problems. However all RRTs suffer from the fundamental limitation of slow convergence in spaces with a narrow channel between the Start and Goal states because of the small probability of randomly generating points inside narrow channels. In this paper an algorithm that attempts to generate points inside narrow channels with a predetermined bias was presented. A point is considered to fall inside a narrow channel if a majority of its neighboring points belong to the infeasible set. The Narrow Channel RRT (**NCRRT**) algorithm proposed in this paper was compared to the standard RRT and its variants on benchmark path planning problems. Simulation results indicate that the **NCRRT** algorithm computes a significantly shorter path in problems with narrow channels. Future work might explore the application of **NCRRT** to practical path planning problems where narrow channels in the configuration space naturally arise such as path planning for mobile robots inside congested buildings [22]. The effects of combining the narrow channel bias heuristic with other heuristics like goal bias can also be explored.